\documentclass[10pt,twocolumn,letterpaper]{article}
\usepackage[pagenumbers]{cvpr} %

\usepackage{amsmath,amsfonts,amsthm,amssymb}
\usepackage{mathtools}
\usepackage{bm}
\usepackage{nicefrac}
\usepackage{microtype}
\usepackage{lipsum}

\usepackage{color,xcolor}
\usepackage{epsfig}
\usepackage{graphicx}

\usepackage{adjustbox}
\usepackage{array}
\usepackage{booktabs}
\usepackage{colortbl}
\usepackage{wrapfig}
\usepackage{hhline}
\usepackage{multirow}
\usepackage{subcaption}
\usepackage[size=small]{caption}
\usepackage{float}

\usepackage{changepage}
\usepackage{extramarks}
\usepackage{fancyhdr}
\usepackage{lastpage}
\usepackage{setspace}
\usepackage{soul}
\usepackage{xspace}

\usepackage{url}

\usepackage{algpseudocode}
\usepackage{algorithmicx}
\usepackage[ruled]{algorithm2e}
\usepackage{enumerate}
\usepackage{enumitem}  %
\usepackage{makecell}
\usepackage{pifont}

\usepackage{framed}

\definecolor{formalbar}{rgb}{0.290,0.325,0.337}
\definecolor{formalshade}{rgb}{0.925,0.941,0.976}

\newcolumntype{L}[1]{>{\raggedright\let\newline\\\arraybackslash\hspace{0pt}}m{#1}}
\newcolumntype{C}[1]{>{\centering\let\newline\\\arraybackslash\hspace{0pt}}m{#1}}
\newcolumntype{R}[1]{>{\raggedleft\let\newline\\\arraybackslash\hspace{0pt}}m{#1}}

\newcommand{\fig}[1]{Fig.~\ref{#1}}

\newcommand{\ignore}[1]{}
\newcommand{\norm}[1]{\lVert#1\rVert}

\DeclareMathAlphabet{\mathbfit}{OML}{cmm}{b}{it}

\makeatletter
\DeclareRobustCommand\onedot{\futurelet\@let@token\@onedot}
\def\@onedot{\ifx\@let@token.\else.\null\fi\xspace}

\def\wrt{w.r.t\onedot}

\def\etal{et al\onedot}

\def\alp{ALP}
\def\alps{ALPs}
\def\supmat{supplementary material}

\makeatother

\definecolor{MyDarkBlue}{rgb}{0,0.08,1}
\definecolor{MyAqua}{rgb}{0,0.7,0.7}
\definecolor{MyDarkGreen}{rgb}{0.02,0.6,0.02}
\definecolor{MyDarkRed}{rgb}{0.8,0.02,0.02}
\definecolor{MyDarkOrange}{rgb}{0.40,0.2,0.02}
\definecolor{MyPurple}{RGB}{111,0,255}
\definecolor{MyRed}{rgb}{1.0,0.0,0.0}
\definecolor{MyGold}{rgb}{0.75,0.6,0.12}
\definecolor{MyDarkgray}{rgb}{0.66, 0.66, 0.66}

\newif\ifdrafting
\draftingtrue %
\ifdrafting
    \newcommand{\jw}[1]{\textcolor{MyDarkGreen}{[Jiajun: #1]}}
    \newcommand{\ky}[1]{\textcolor{red}{[Koven: #1]}}
    \newcommand{\ds}[1]{{\leavevmode\color[rgb]{0.8,0.2,0}[Deqing: #1]}}
    \newcommand{\cih}[1]{{\textcolor{MyAqua}{[Charles: #1]}}}
    \newcommand{\samirag}[1]{\textcolor{blue}{[Samir: #1]}}
    
\else
    \newcommand{\ds}[1]{}
    \newcommand{\cih}[1]{}
    \newcommand{\jw}[1]{}
    \newcommand{\ky}[1]{}
    \newcommand{\samirag}[1]{}
\fi

\SetKwFor{For}{for }{}{}

\DeclareMathOperator{\Loss}{\mathcal{L}}

\newcommand{\myparagraph}[1]{\vspace{0.1cm}\noindent\textbf{#1}}

\newenvironment{myitemize}
{ \vspace{-0.25cm}
\begin{itemize}
    \setlength{\itemsep}{0pt}
    \setlength{\parskip}{0pt}
    \setlength{\parsep}{0pt}     }
{ \end{itemize}
\vspace{-0.1cm}}

\usepackage[pagebackref,breaklinks,colorlinks]{hyperref}

\usepackage[capitalize]{cleveref}
\crefname{section}{Sec.}{Secs.}
\Crefname{section}{Section}{Sections}
\Crefname{table}{Table}{Tables}
\crefname{table}{Tab.}{Tabs.}

\def\cvprPaperID{3978} %
\def\confName{CVPR}
\def\confYear{2023}

\begin{document}

\title{Accidental Light Probes}

\author{Hong-Xing Yu\textsuperscript{1}  \hspace{5mm} %
Samir Agarwala\textsuperscript{1} \hspace{5mm}
Charles Herrmann\textsuperscript{2} \hspace{5mm}
Richard Szeliski\textsuperscript{2} \hspace{5mm}
Noah Snavely\textsuperscript{2} 
\vspace{0.15cm}\\
Jiajun Wu\textsuperscript{1} \hspace{5mm}
Deqing Sun\textsuperscript{2} \hspace{5mm}
\vspace{0.25cm}\\
\textsuperscript{1}Stanford University  \hspace{10mm}
\textsuperscript{2}Google Research
}

\maketitle

\begin{abstract}
Recovering lighting in a scene from a single image is a fundamental problem in computer vision. While a mirror ball light probe can capture omnidirectional lighting, light probes are generally unavailable in everyday images. In this work, we study recovering lighting from \emph{accidental} light probes (ALPs)---common, shiny objects like Coke cans, which often accidentally appear in daily scenes. We propose a physically-based approach to model ALPs and estimate lighting from their appearances in single images. The main idea is to model the appearance of ALPs by photogrammetrically principled shading and to invert this process via differentiable rendering to recover incidental illumination. 
We demonstrate that we can put an ALP into a scene to allow high-fidelity lighting estimation. Our model can also recover lighting for existing images that happen to contain an ALP\footnote{Project website: \url{https://kovenyu.com/ALP}}.
\vspace{-0.5cm}
\end{abstract}

\vspace{-10pt}
\begin{flushright}
\emph{I'd rather be Shiny.} ---  Tamatoa from Moana, 2016
\end{flushright}

\section{Introduction}

\begin{figure}
    \centering
    \includegraphics[width=0.95\linewidth]{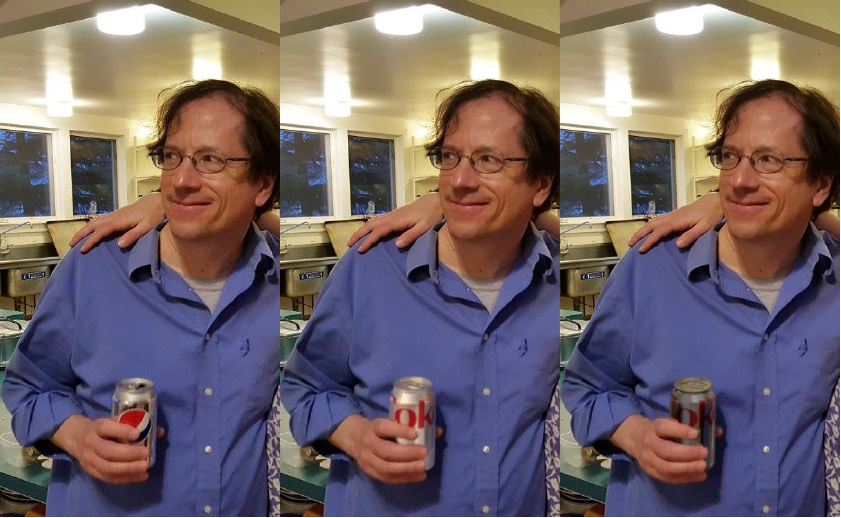}
    \caption{(Left) From an image that has an accidental light probe (a Diet Pepsi can), we insert a virtual object (a Diet Coke can) with estimated lighting using the accidental light probe (Middle), and using estimated lighting from a recent state-of-the-art lighting estimation method~\cite{wang2022stylelight} (Right). Note how our method better relights the inserted can to produce an appearance consistent with the environment (e.g., the highlight reflection and overall intensity).}
    \label{fig:teaser}
\end{figure}

Traditionally, scene lighting has been captured through the use of light probes, typically a chromium mirror ball; their shape (perfect sphere) and material (perfect mirror) allow for a perfect measurement of all light that intersects the probe. Unfortunately, perfect light probes rarely appear in everyday photos, and it is unusual for people to carry them around to place in scenes. Fortunately, many everyday objects share the desired properties of light probes: Coke cans, rings, and thermos bottles are shiny (high reflectance) and curved (have a variety of surface normals). These objects can reveal a significant amount of information about the scene lighting, and can be seen as imperfect ``accidental'' light probes (e.g., the Diet Pepsi in Figure~\ref{fig:teaser}). %
Unlike perfect light probes, they can easily be found in casual photos or acquired and placed in a scene. In this paper, we explore using such everyday, shiny, curved objects as Accidental Light Probes (ALPs) to estimate lighting from a single image.

In general, recovering scene illumination from a single view is fundamental for many computer vision applications such as virtual object insertion~\cite{debevec1998rendering}, relighting~\cite{sun2019single}, and photorealistic data augmentation~\cite{wang2022neural}. Yet, it remains an open problem primarily due to its highly ill-posed nature. Images are formed through a complex interaction between geometry, material, and lighting~\cite{kajiya1986rendering}, and without precise prior knowledge of a scene's geometry or materials, lighting estimation is extremely under-constrained. 
For example, scenes that consist primarily of matte materials reveal little information about lighting, since diffuse surfaces behave like low-pass filters on lighting during the shading process~\cite{ramamoorthi2001signal}, eliminating high-frequency lighting information. 
To compensate for the missing information, the computer vision community has explored using deep learning to extract data-driven priors for lighting estimation~\cite{srinivasan2020lighthouse,garon2019fast}. However, these methods generally do not leverage physical measurements to address these ambiguities, yet physical measurements can offer
substantial benefits in such an ill-posed setting.

For images with ALPs, we propose a physically-based modeling approach for lighting estimation. The main idea is to model the ALP appearance using physically-based shading and to invert this process to estimate lighting. This inversion process involves taking an input image, estimating the ALP's 6D pose and scale, and then using the object's surface geometry and material to infer lighting.
Compared to purely data-driven learning approaches that rely on diverse, high-quality lighting datasets, which are hard to acquire, our physically-based approach generalizes to different indoor and outdoor scenes.

To evaluate this technique, we collect a test set of real images, where we put ALPs in daily scenes and show that our approach can estimate high-fidelity lighting. We also demonstrate lighting estimation and object insertion based on existing images (Figure~\ref{fig:teaser}).

In summary, we make the following three contributions:
\vspace{0.1cm}
\begin{myitemize}
    \item We propose the concept of \emph{accidental} light probes (ALPs), which can provide strong lighting cues in everyday scenes and casual photos.
    \item We develop a physically-based approach for lighting estimation for images with an ALP and show improved visual performance compared to existing light estimation techniques.
    \item We collect a dataset of ALPs and a dataset of images with ALPs and light probes in both indoor and outdoor scenes. We demonstrate that our physically-based model outperforms existing methods on these datasets.
\end{myitemize}

\section{Related Work}

\myparagraph{Lighting estimation.} Traditional light probes capture omnidirectional lighting~\cite{debevec1998rendering} but are usually absent in existing images. Researchers have used everyday objects like human faces~\cite{legendre2020learning,calian2018faces,sun2019single,yi2018faces, knorr2014real} and eyes~\cite{nishino2004eyes} to estimate lighting in portrait images. In contrast, we target images with high-reflectance objects. Other research focuses on lighting estimation from known, non-reflective objects. 
Weber \etal~\cite{weber2018learning} and Park \etal~\cite{park2020physically} learn to regress illumination directly from homogeneous-material objects, while \etal~\cite{wei2020object} extend this to spatially-varying materials.
These methods require large, diverse lighting data to generalize. Some approaches use RGBD video~\cite{park2020seeing,richter2016instant} to estimate scene lighting, but we focus on lighting estimation from a single RGB image.

In addition to object-based lighting estimation, another popular line of work focuses on learning lighting estimation directly from images of scenes~\cite{gardner2019deep,garon2019fast,zhan2021emlight,zhan2021sparse,gardner2017learning,song2019neural,srinivasan2020lighthouse}. Many of these methods rely heavily on supervised training on synthetic data. As a result, 
they are sensitive to domain shifts between training and test data and, in particular, suffer from a synthetic-to-real domain gap. In contrast, our approach is based on physically principled modeling and is not vulnerable to this issue.

\myparagraph{Inverse rendering.} 
Our approach is closely related to inverse rendering methods that aim to jointly recover geometry, material, and lighting from images. Recent work in this area uses multi-view observations of an object with known camera poses to recover scene lighting and object properties~\cite{munkberg2022extracting, hasselgren2022shape}. These methods jointly optimize geometry, material, and lighting and generalize to diverse scene settings. However, in single-view settings, the optimization problem for inverse rendering is highly ill-posed, and these methods often produce degenerate solutions.

Learning-based inverse rendering techniques have also gained popularity in material and geometry estimation tasks \cite{StyleGAN3D, neuralSengupta19, li2020inverse, wang2021learning, yu19inverserendernet}. These methods include differential rendering as part of their training pipeline and can learn priors to model geometry and materials of scenes and objects.  However, they are limited in their ability to generalize to a diverse set of scenes.

\myparagraph{Material reconstruction.}
Material modeling and reconstruction have a long history in computer vision and graphics. Early papers~\cite{torrance1967theory,blinn1977models, goral1984modeling} developed early analytical models of material reflection based on general experimental observations. More recent works~\cite{dana1999reflectance, jensen2001practical, lensch2003image} have attempted to directly solve for a general bidirectional reflectance distribution function (BRDF), which analytically defines how light is reflected at a given point on an object's surface; however, many of these techniques fail for highly specular or curved objects. For example, traditional BRDF acquisition~\cite{marschner2000image,dupuy2018adaptive} requires a gonioreflectometer, which tries to precisely measure reflectance at different angles. This machine typically runs on flat objects and struggles on curved objects like Diet Coke cans. Modern approaches~\cite{zollhofer2018state} use RGBD sensors and joint optimization on differentiably rendered objects and multi-view images~\cite{hasselgren2022shape, munkberg2022extracting, zhang2022iron}; our approach builds upon differentiable rendering to optimize material reconstruction and adapts them for ALPs.
\section{Approach}

\begin{figure*}[t]
\centering
\includegraphics[width=1\linewidth]{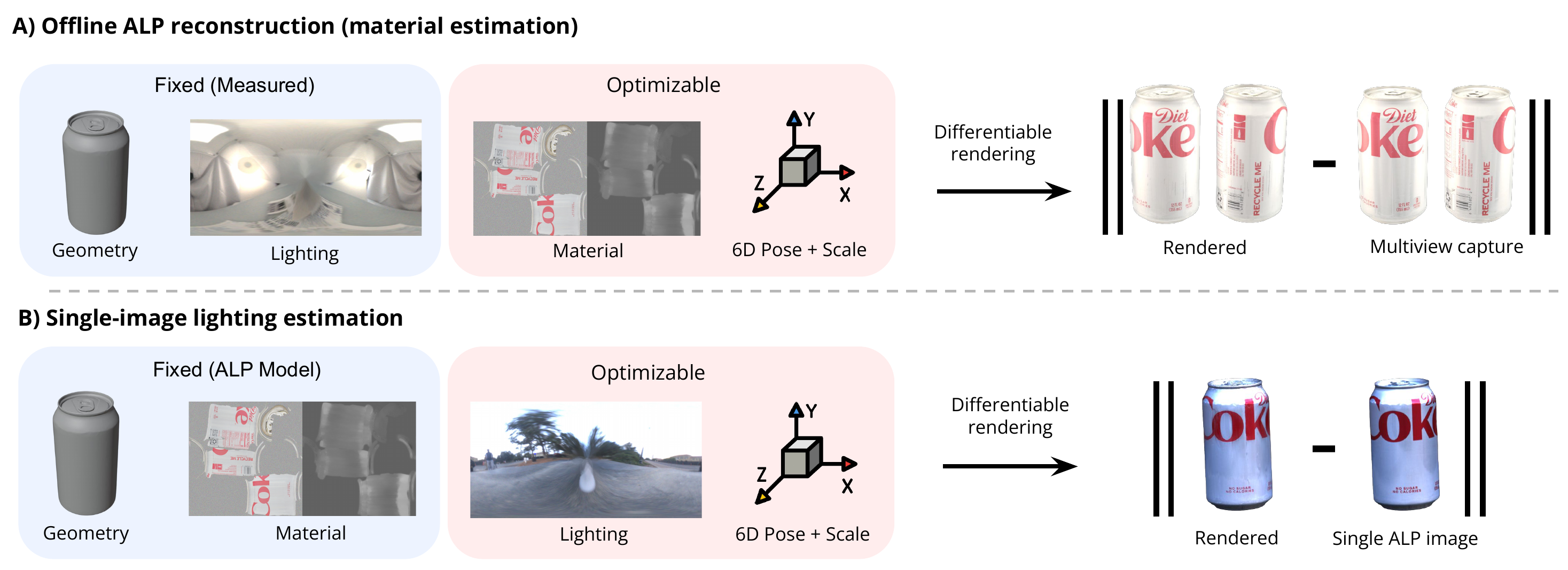}
\vspace{-0.15in}
\captionof{figure}{Our physically-based approach to lighting estimation consists of (A) offline Accidental Light Probe (ALP) reconstruction and (B) inference-time single-image lighting estimation. For (A), we use a capture-optimization hybrid method to reconstruct the ALP model with high fidelity. For (B), we formulate lighting estimation as a joint optimization of scale, 6D pose and environment lighting.
}
\label{fig:overview}
\end{figure*}

Accidental Light Probes (ALPs) are daily metallic shiny objects, such as a soda can, a thermoflask, or a ring.
Given a single image containing an ALP, we aim to recover the incidental illumination by inverting physically-based rendering, as shown in \fig{fig:overview}. Our main idea is that we can first acquire the shape and reconstruct the spatially-varying BRDF of the ALP offline (\fig{fig:overview} top), and then optimize incidental lighting as well as the 6D pose of the ALP (\fig{fig:overview} bottom).

\subsection{Formulation}\label{formulation}
Our goal is to estimate high-fidelity lighting from the appearance of an ALP in a single image. We approach this goal through the perspective of inverse rendering, where the forward process is described by the rendering equation~\cite{kajiya1986rendering}:
\begin{align}
    L(\omega_o) = \int_H L_i(\omega_i) f(\omega_i, \omega_o) (n\cdot \omega_i) d\omega_i,
    \label{eqn:render}
\end{align}
where $L(\omega_o)$ is the outgoing radiance to direction $\omega_o$ (corresponding to pixel intensity), $L_i(\omega_i)$ is incidental radiance from direction $\omega_i$ (lighting), $f$ is the bidirectional reflectance distribution function (BRDF) at the surface location (material), $n$ is the normal direction (geometry), and $H$ is the upper hemisphere along the normal. 
Recovering lighting by inverting Eqn~\ref{eqn:render} is a highly ill-posed problem, as infinitely many combinations of geometry, material, and lighting can generate the same appearance in the image. Fortunately, for ALPs, we can pre-acquire prior physical knowledge of their shapes and materials  as they are everyday objects. Thus, we can reduce the full inverse rendering problem to a joint estimation of 6D ALP pose and lighting, which is relatively more constrained and tractable:
\begin{align}
    \min_{\pi, L_i}  \Loss \big ( I_{\text{render}}(\pi, L_i | f, S), I_{\text{ref}} \big ),
    \label{eqn:joint}
\end{align}
where $I_\text{render}$ is generated by a differentiable renderer that takes the shape $S$ (represented by a mesh), the 6D pose of the ALP $\pi$, the spatially-varying BRDF $f$, and the environment lighting $L_i$ as inputs. $I_\text{ref}$ denotes the observed single image. $\Loss$ denotes an image-space loss that we define in Section~\ref{sec:light}.
Our physically-based formulation entails the high-fidelity acquisition of shape and spatially-varying material of the ALP, as well as a robust single-view joint optimization algorithm. We show an overview in \fig{fig:overview} and elaborate the components in Section~\ref{sec:acquire} and Section~\ref{sec:light}, respectively.

\myparagraph{Shading model.} We adopt physically-based rendering (PBR)~\cite{pharr2016physically} due to its principled photogrammetry and radiometry. Specifically, we consider metallic materials as they have little diffuse reflection. Diffuse reflection is undesirable as it behaves like a low-pass filter of lighting in the shading process~\cite{ramamoorthi2001signal}, eliminating the physically recoverable lighting information. To model metallic material, we use a microfacet model~\cite{torrance1967theory} with a GGX distribution~\cite{walter2007microfacet}:
\begin{align}
    f(\omega_i, \omega_o) = \frac{D\cdot F\cdot G}{4(n\cdot\omega_i)(n\cdot\omega_o)},
    \label{eqn:microfacet}
\end{align}
where $D$ is the GGX normal distribution~\cite{walter2007microfacet}, $F$ is the Fresnel reflection, and $G$ is the geometric attenuation. We adopt Disney's parameterization~\cite{burley2012physically}, where the metallic material is modeled by its specular albedo $A$ and roughness $r$. Specifically, the specular albedo $A$ is used to model Fresnel reflection by Schlick's approximation~\cite{schlick1994inexpensive} $F=A+(1-A)(1-|h\cdot\omega_o|)^5$, where $h=\frac{\omega_i+\omega_o}{|\omega_i+\omega_o|}$ denotes the half vector. The roughness $r$ controls the shape of the specular reflection lobe via the micro-normal distribution $D=\frac{r^4}{\pi (|n\cdot h|^2(r^4-1)+1)^2}$ and the geometric attenuation $G=\frac{2|n\cdot \omega_i|}{|n\cdot \omega_i|+\sqrt{r^4+(1-r^4)|n\cdot \omega_i|^2}}\cdot\frac{2|n\cdot \omega_o|}{|n\cdot \omega_o|+\sqrt{r^4+(1-r^4)|n\cdot \omega_o|^2}}$.

\myparagraph{Lighting model.} To recover lighting for arbitrary conditions, we use an environment map to represent omnidirectional lighting and adopt image-based lighting for shading each pixel. For efficiency, we only consider direct lighting and use a differentiable rasterizer with deferred shading~\cite{laine2020modular} to render $I_\text{render}$. This is inaccurate for concave objects with self-occlusion and self-reflections. To mitigate this without expensive global illumination, we include a soft visibility term to Eqn~\ref{eqn:render} to approximate it such that the shading output is modulated as $vL(\omega_o)$, where $v$ denotes the soft visibility that is optimized and treated as a surface texture.

\subsection{Reconstructing \alps}\label{sec:acquire}
\begin{figure}[t]
\centering
\includegraphics[width=\linewidth]{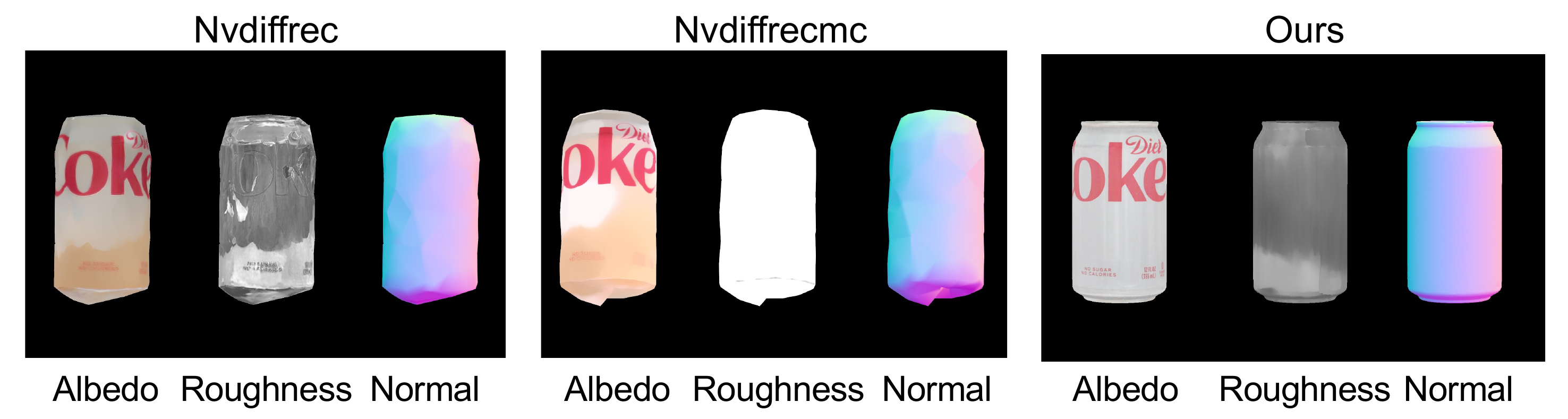}
\vspace{-0.15in}
\captionof{figure}{Visual comparison of ALP reconstruction from state-of-the-art optimization-based inverse rendering methods~\cite{hasselgren2022shape,munkberg2022extracting} versus our hybrid method. Recent inverse rendering methods struggle on real textured metallic objects.}
\label{fig:capture}
\end{figure}

Recovering high-fidelity lighting by physically-based inverse rendering requires high-quality geometry and material reconstruction of the ALPs. While existing state-of-the-art inverse rendering methods can jointly optimize for geometry, material, and lighting from dense multi-view images~\cite{hasselgren2022shape,zhang2022iron,munkberg2022extracting}, they still struggle for real metallic objects under arbitrary lighting due to high specularity (\fig{fig:capture}). Moreover, several challenges exist when the goal is not view synthesis but photogrammetrically correct reconstruction. For highly specular objects such as metallic ones, the reflected lights from the near field can lead to environment baking, as it breaks the distant light assumption (we show an example of the environment-baked material reconstruction in the \supmat). The color ambiguity of material albedo and lighting is also not resolved. In addition, the geometry reconstruction quality heavily relies on the quality of object silhouettes in multi-view images.

\begin{figure}[t]
\centering
\includegraphics[width=1\linewidth]{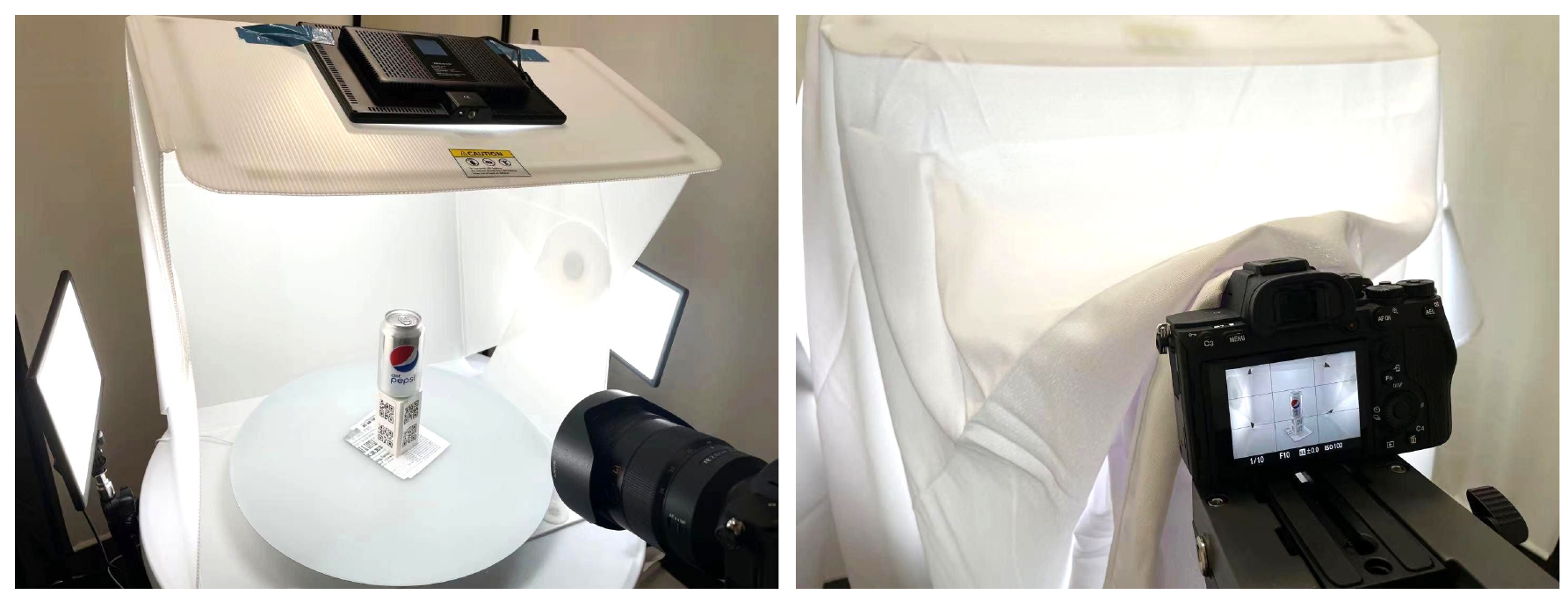}
\vspace{-0.15in}
\captionof{figure}{(Left) We use a light box with controllable lighting for our capture. To mitigate near-field reflections, we leverage a thin stand to support the object. (Right) To minimize environmental changes due to camera and photographer movement, we cover the lightbox with a cloth and use a turntable for multi-view capture.}
\label{fig:capture_setup}
\end{figure}

To overcome these challenges, we reconstruct \alps~by a hybrid method. First, we use a light box with a turntable to control environment lighting for multi-view capture, and using a thin supporting stand to alleviate near-field reflections (setup shown in \fig{fig:capture_setup}) and environment baking. Second, instead of optimizing the incidental lighting to the \alp~under capture, we record it by a calibrated light probe to remove the color ambiguity between material and lighting. And third, we provide a high-quality shape using a range scanner~\cite{scanner} to reduce the geometry reconstruction down to 6D pose and size fitting. 
Thus, as demonstrated in the top row of Figure~\ref{fig:overview}, our ALP reconstruction is cast as an optimization for its spatially-varying material and shape fitting: 
\begin{align}
    \min_{\pi, \alpha, f}  \sum_{\{I_\text{capture}\}} \Loss \big ( I_{\text{render}}(\pi, \alpha, f | L_i, S), I_{\text{capture}} \big ),
    \label{eqn:reconstruction}
\end{align}
where $\pi$ and $\alpha$ are the 6D pose and size to fit the shape $S$ to multi-view camera coordinate frame solved by COLMAP~\cite{schonberger2016structure}, and $f$ is the material parameterized by spatially-varying albedo $A$ and roughness $r$. We show the reconstruction of a Coke can in \fig{fig:capture}. We include the optimization and loss details in the \supmat.

\subsection{Single-View Physically-Based Light Estimation}\label{sec:light}
Given an image containing an \alp, we first extract an object segmentation mask for the \alp~by manually cropping the image and then using an off-the-shelf foreground segmentation tool~\cite{rmbg}; however, this could alternately be obtained by object detection~\cite{carion2020end} with salient object segmentation~\cite{qin2020u2} or semantic segmentation~\cite{strudel2021segmenter}. We then retrieve the appropriate \alp~ model, containing  its reflectance and geometric information. 
Yet, even given the \alp's 3D model and 2D segmentation in the input image, accurately aligning these two elements is still challenging. Traditional feature point detection and Perspective-n-Point methods do not work on textureless objects such as rings and thermoflasks. Additionally, modern learning-based single-view pose estimation methods~\cite{wang2019normalized,li2018deepim,xiang2017posecnn} require diverse, realistic lighting to synthesize training data and do not generalize well outside the training distribution. 

Therefore, we formulate the lighting estimation and pose estimation as a joint estimation problem in Eqn~\ref{eqn:joint}, and we solve it via a differentiable rendering-based optimization which is generalizable to arbitrary scenes for both textured and textureless objects (see the bottom of Figure~\ref{fig:overview}). Here we need a joint estimation as the appearance of a specular object (and thus the differentiable rendering gradient signals) is highly dependent on both the object pose and the environment lighting. We use Monte Carlo ray tracing with Visible Normal Distribution Function (VNDF) importance sampling~\cite{heitz2018sampling} for unbiased shading.

\myparagraph{Losses and regularizations.} \label{sec:losses}
Our loss function used in Eqn~\ref{eqn:joint} is given by:
\begin{align}
    \Loss = \Loss_\text{RGB} + \Loss_\text{mask} + \lambda_1\Loss_\text{pose-reg} + \lambda_2\Loss_\text{light-reg},
\end{align}
where $\Loss_\text{RGB}$ denotes a $L_1$ loss on RGB images, $\Loss_\text{mask}$ denotes a combination of a $L_1$ loss and a Chamfer loss on masks~\cite{balan2007detailed}, where the mask is given by the differentiable rasterizer~\cite{laine2020modular}. $\Loss_\text{pose-reg}$ and $\Loss_\text{light-reg}$ denote a pose regularization and a lighting regularization with their weights $\lambda_1$ and $\lambda_2$, respectively.

Without multi-view constraints, the joint optimization problem has multiple local minima for the 6D pose;  thus, we introduce a pose regularization and a lighting regularization. The pose regularization is given by:
\begin{align}
    \Loss_\text{pose-reg} = \norm{B(M_\text{render}) - B(M_\text{ref})}_2^2 + \norm{q - q_\text{ref}}_2^2,
    \label{eqn:pose_reg}
\end{align}
where $M_\text{render}$ is the rendered mask, $B(M_\text{render})$ denotes the pixel-space barycenter of the mask, $q$ denotes the quaternion representation of the ALP orientation, and $q_\text{ref}$ denotes a common orientation (we use a front-facing canonical orientation obtained by aligning principal axes). The barycenter term prevents vanished gradient due to non-overlapping pose initialization, and the orientation term prevents hard-to-escape local minima like upside-down cans. 
We decay the weight of the pose regularization to zero through optimization.
In addition, to further address local minima in 6D poses, we use multiple (4 in our experiments) orientation initialization and we keep the one with the highest re-rendering PSNR.

To accurately estimate omnidirectional lighting by inverting Eqn~\ref{eqn:render}, we need to evaluate the Monte Carlo integral interval densely over light rays coming from all directions. However, from a single view, an \alp~often only covers a limited subset of normal directions compared to a perfect sphere. Thus, light rays coming from a certain subset of directions contribute little to the appearance of the \alp. These directions are then under-sampled, and the lighting estimation for them is less informed and unconfident.

To mitigate this, we introduce a lighting smoothness regularization which ``fills in'' the less confident regions in the environment map by propagating the confident information from nearby directions. The regularization is given by:
\begin{align}
    \Loss_\text{light-reg} = \norm{L_i(\omega) - L_i(\omega+\Delta\omega)}_1,
\end{align}
where $\Delta\omega$ denotes a small deviation of a solid angle sampled from a normal distribution, and $\omega$ is sampled uniformly in all solid angles.
Note that in addition to propagating confident lighting estimates, the lighting regularization also helps improve pose estimation, since many pose estimation errors come from trying to fix mistakes in high-frequency~\cite{hartung2002distinguishing} lighting changes, which light regularization alleviates.

\section{Experiments}

\begin{figure}[t]
\centering
\includegraphics[width=1\linewidth]{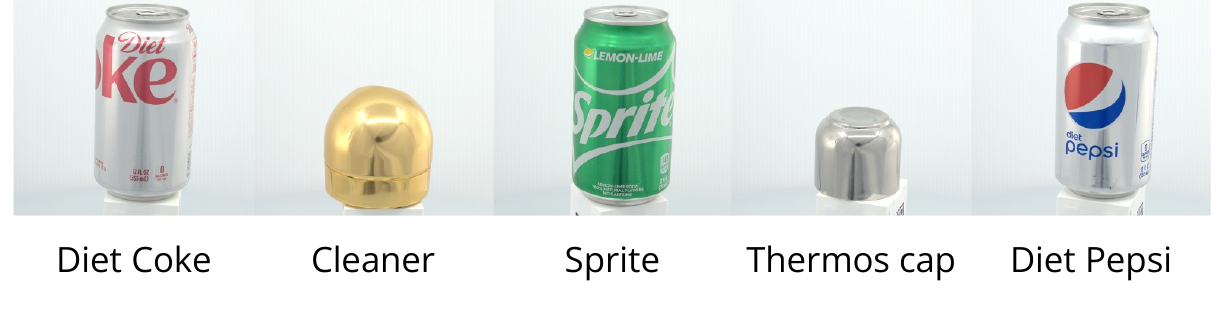}
\vspace{-0.15in}
\captionof{figure}{Close up of our \alp~dataset.
}
\label{fig:closeup}
\end{figure}

\subsection{Setup}
\noindent\textbf{Accidental Light Probes Dataset.}
We acquire 5 common accidental light probes that have different shapes or spatially-varying BRDFs, including 3 soda cans (diet Coke, diet Pepsi, and Sprite), a thermos cap, and a solder tip cleaner. We show example images in Figure \ref{fig:closeup}.

\begin{figure*}[t]
\centering
\includegraphics[width=\linewidth]{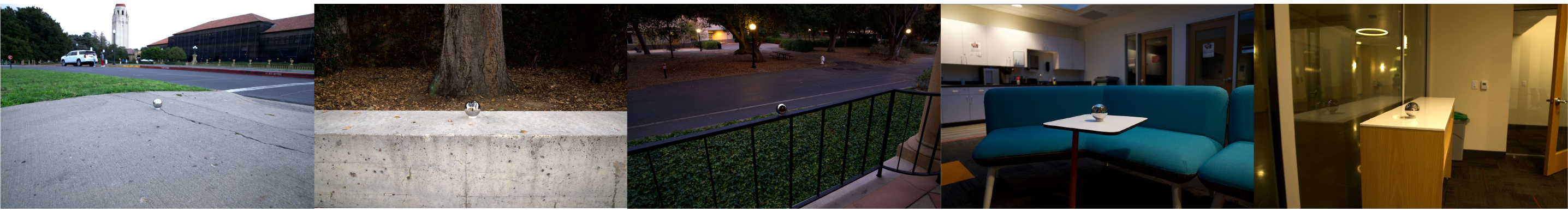}
\vspace{-0.15in}
\caption{Examples of our collected dataset for evaluating lighting estimation under different illumination conditions, including indoor and outdoor scenes at daytime and nighttime.}
\label{fig:dataset_examples}
\end{figure*}

\myparagraph{Evaluation Dataset.}
We collect a dataset of 10 indoor scenes and 13 outdoor scenes. The indoor and outdoor scenes are taken at different points of time, such as day and night, at different locations. We show examples in Figure~\ref{fig:dataset_examples}. We place each of our ALPs in the scenes and capture HDR images of the ALPs. We also capture ground-truth lighting by a chromium ball (a perfect light probe).

\myparagraph{Baselines.}
We compare our method to several state-of-the-art lighting estimation methods~\cite{garon2019fast, gardner2019deep, wang2022stylelight}. Unlike our method, all of these techniques utilize deep learning. Since ~\cite{garon2019fast, gardner2019deep} do not have publicly available models, we asked their authors to run inference on our dataset. 

\begin{figure*}[t]
\centering
\includegraphics[width=1\linewidth]{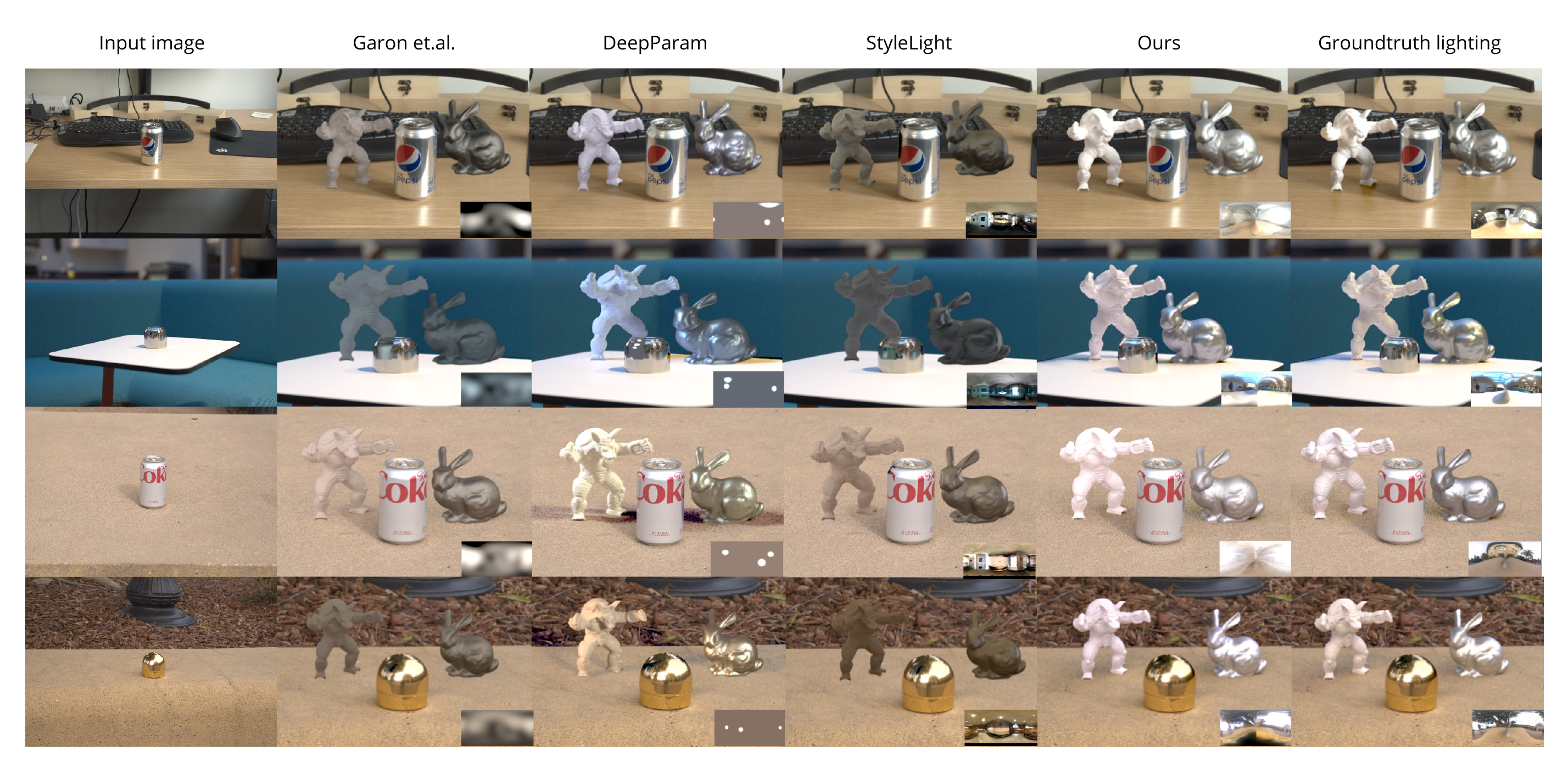}
\vspace{-0.15in}
\captionof{figure}{Object insertion results on our test scenes for both indoor (first two rows) and outdoor (last two rows). We compare to Garon et al.~\cite{garon2019fast}, Deep parametric lighting~\cite{gardner2019deep}, and StyleLight~\cite{wang2022stylelight}. We center-crop the result images for better visualization.
}
\label{fig:obj_insertion}
\end{figure*}

\begin{table*}[t]
    \centering
    \small
    \begin{tabular}{lcccccccccccc}
        \toprule
        \multirow{4}{*}{Method}  & \multicolumn{6}{c}{\textbf{Indoor}} & \multicolumn{6}{c}{\textbf{Outdoor}} \\
        \cmidrule(lr){2-7}\cmidrule(lr){8-13}
        & \multicolumn{3}{c}{\emph{Angular Error}$\downarrow$} &
        \multicolumn{3}{c}{\emph{Scale-invariant RMSE}$\downarrow$}&
        \multicolumn{3}{c}{\emph{Angular Error}$\downarrow$} &
        \multicolumn{3}{c}{\emph{Scale-invariant RMSE}$\downarrow$}
        \\
        \cmidrule(lr){2-4}\cmidrule(lr){5-7}\cmidrule(lr){8-10}\cmidrule(lr){11-13}
          &  Mirror & Shiny & Diffuse &  Mirror & Shiny & Diffuse &  Mirror & Shiny & Diffuse &  Mirror & Shiny & Diffuse \\
        \midrule
        StyleLight~\cite{wang2022stylelight} & 12.572 & 7.700 & 5.949 & 3.087 & 0.837 & 0.264 & 15.088 & 9.830 & 8.539 & 1.867 & 0.918 & 0.294   \\
        Deep Param.~\cite{gardner2019deep} & 7.204 & 6.252 & 6.166 & 3.137 & 0.958 & 0.287 & 8.803 & 7.228 & 6.525 & 1.963 & 1.056 & 0.305 \\
        Garon \etal~\cite{garon2019fast} & 9.403 & 8.215 & 6.626 & 3.030 & 0.754 & 0.207 & 8.062 & 6.873 & 6.118 & 1.706 & 0.766 & 0.237 \\
        \midrule
        Cleaner & 5.682 & 4.550 & 3.965 & \textbf{2.204} & \textbf{0.252} & 0.073 & 6.395 & 4.920 & 5.155 & \textbf{1.081} & 0.245 & 0.101    \\
        Diet Coke & 4.733 & 3.405 & 3.067 & 2.901 & 0.550 & 0.101 & 6.011 & 3.877 & 2.587 & 1.460 & 0.501 & 0.136  \\
        Diet Pepsi & 3.972 & 2.712 & 2.190 & 2.726 & 0.408 & 0.064 & 4.890 & 2.830 & \textbf{1.472} & 1.352 & 0.396 & 0.108 \\
        Sprite & 5.952 & 4.445 & 3.767 & 2.913 & 0.556 & 0.112 & 7.023 & 4.923 & 3.892 & 1.468 & 0.513 & 0.154   \\
        Thermos cap & \textbf{3.744} & \textbf{2.080} & \textbf{1.622} & 2.555 & 0.288 & \textbf{0.057} &  \textbf{3.965} & \textbf{2.159} & 1.516 & 1.092 & \textbf{0.217} & \textbf{0.053}\\
        \bottomrule
    \end{tabular}
    \vspace{-0.2cm}
    \caption{Comparison to state-of-the-art single image lighting estimation methods: StyleLight~\cite{wang2022stylelight}, Deep Parametric~\cite{gardner2019deep} and Garon et al~\cite{garon2019fast}. We evaluate them using relighting on different materials.}
    \vspace{-0.2cm}
    \label{tbl:baseline_comp}
\end{table*}

\begin{figure*}[t]
\centering
\includegraphics[width=\linewidth]{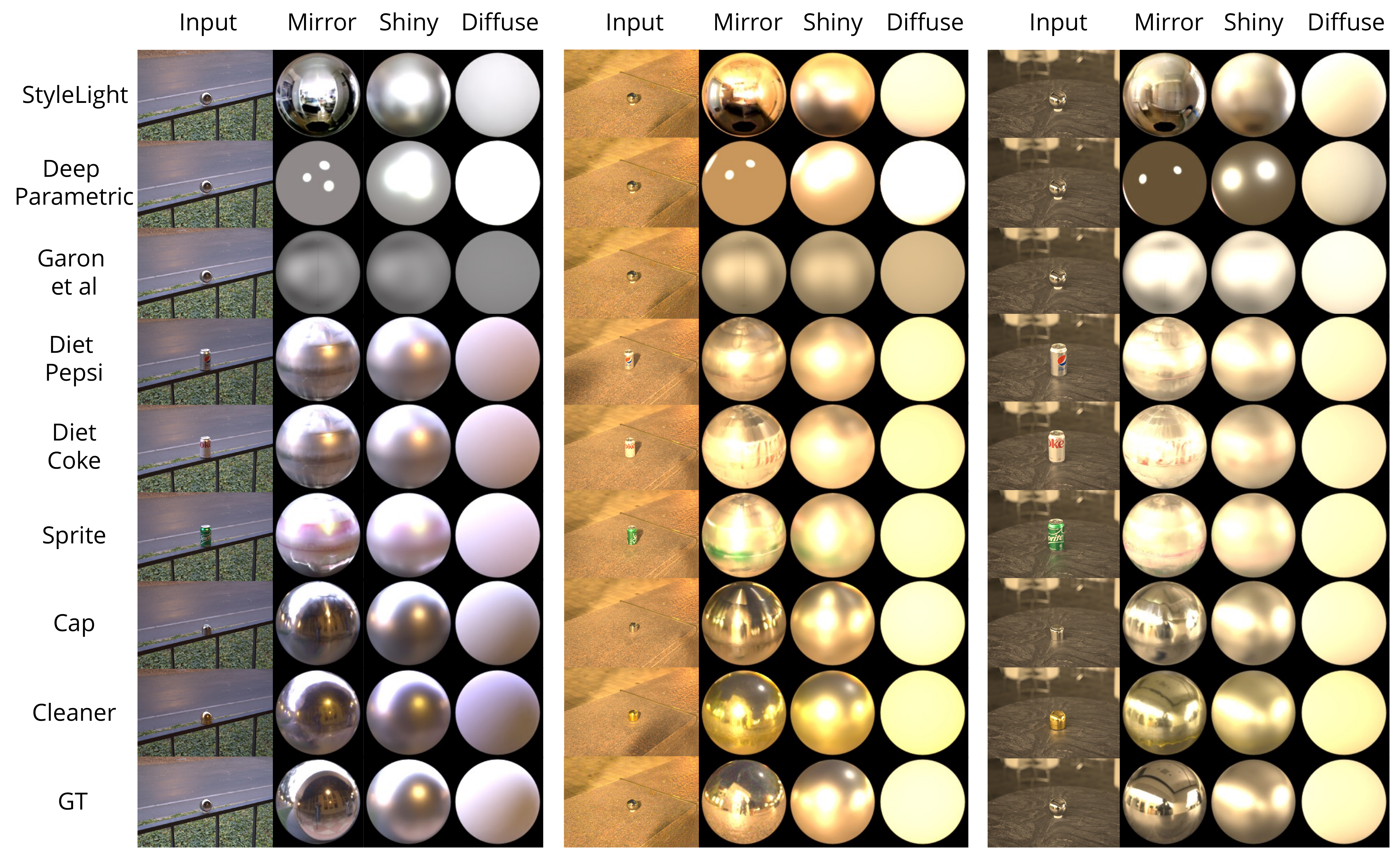}
\vspace{-0.2in}
\caption{Qualitative comparison of relighting results in outdoor (left and center) and indoor (right) scenes. We compare our approach to StyleLight~\cite{wang2022stylelight}, Deep Parametric~\cite{gardner2019deep} and Garon et al~\cite{garon2019fast} on relighting mirror, shiny and diffuse spheres.}
\label{fig:qualitative_comp}
\end{figure*}

\subsection{Comparison to Baseline Methods}

\noindent\textbf{Qualitative Results.} 
For all object insertion comparisons, we compute an environment map either through an ALP with our proposed method or by running the other baselines on a single image of the scene. Note that for the baselines, we use the image with the perfect light probe as input; this should provide a slight advantage to these techniques since the image with the perfect light probe contains the most information regarding scene lighting.

In Figure~\ref{fig:obj_insertion}, we insert various objects into the scene and relight them using the computed environment maps; we then qualitatively compare the results. We demonstrate that our computed environment map produces significantly more accurate and compelling results than other single-image lighting estimation methods. In particular, note that our method is the only approach that can recover the overall tone of the lighting: other methods are either too yellow or gray. 

In Figure~\ref{fig:qualitative_comp}, we show relighting results on perfect spheres of various finishes from all methods and ALPs on both indoor and outdoor scenes. Only our technique produces results similar to the ground truth for mirror finishes. We note that for all the three soda cans, the relighting on mirror spheres are slightly blurry, since their materials are much rougher than perfect mirror, which behaves as low-pass filters of lighting in the shading process~\cite{ramamoorthi2001signal}. We also note that for Sprite and diet Coke, there is some texture color baking in the recovered lighting due to imperfectly aligned 6D poses, which lead to high-frequency lighting artifacts to compensate the pixel-space misalignment. Our lighting regularization mitigates this type of artifacts, yet a highly robust algorithm remains as future work.

\myparagraph{Quantitative Results.} 
In Table~\ref{tbl:baseline_comp}, we report quantitative results on relighting perfect spheres with various representative materials (mirror, shiny, diffuse). Similar to~\cite{wang2022stylelight,legendre2019deeplight}, we compute angular error\cite{finlayson2016reproduction} and scale-invariant RMSE~\cite{grosse2009ground} to compare the relighted spheres from each technique to the ground truth relighting.

Quantitatively, for the relighting task, our method, applied to any of the ALPs, significantly outperforms the baselines. In particular, \wrt angular error, the Thermos cap provides a 3 to 4 times improvement over the best baseline. %

\begin{table}[t]
    \centering
    \small
    \begin{tabular}{lccc}
        \toprule
        Method  & Mirror & Shiny & Diffuse \\
        \midrule
        Nvdiffrec~\cite{munkberg2022extracting}  & 6.99 & 5.06 & 3.59 \\
        Nvdiffrecmc~\cite{hasselgren2022shape}&  6.55 & 4.60 & 3.84 \\
        ALP (Ours)  & \textbf{5.46} & \textbf{3.67} & \textbf{2.80} \\
        \bottomrule
    \end{tabular}
    \caption{Evaluation on our ALP model acquisition for a Diet Coke can using our lightbox setup. We compare our acquisition method to Nvdiffrec~\cite{munkberg2022extracting} and Nvdiffrecmc~\cite{hasselgren2022shape}. We use the same lighting estimation approach for compared methods and report average angular error across all test scenes.}
    \label{tbl:abl_config}
\end{table}

\begin{table}[t]
    \centering
    \small
    \begin{tabular}{lccc}
        \toprule
        Method  &  Mirror & Shiny & Diffuse \\
        \midrule
        Silhouette loss~\cite{balan2007detailed} & 6.812 & 4.976 & 3.919 \\
        Ours w/o joint optimization & 5.401 & 3.726 & 3.044  \\
        Ours w/o pose regularization & 5.962 & 4.180 & 3.338\\
        Ours w/o light regularization & 6.032 & 3.647 & 2.954 \\
        Ours & \textbf{5.291} & \textbf{3.610} & \textbf{2.923} \\
        \bottomrule
    \end{tabular}
    \caption{Ablation study on our joint pose-lighting optimization. We compare to a baseline that uses a silhouette loss and a Chamfer loss~\cite{balan2007detailed}, and variants of our approach. We show angular errors averaged on all test scenes.}
    \label{tbl:abl_pose}
\end{table}

\subsection{Analysis}
\myparagraph{Capture Setup.} We also analyze the quality of our reconstruction compared to two recent multi-view inverse rendering methods, Nvdiffrec~\cite{munkberg2022extracting} and Nvdiffrecmc~\cite{hasselgren2022shape} using our lightbox capture setups. Table~\ref{tbl:abl_config} shows the results of using our lighting estimation pipeline with various reconstructions of a Diet Coke can. Our reconstruction performs the best and leads to a decrease of angular error by 20\% or more. Both Nvdiffrec and Nvdiffrecmc are normally applied to multiview casual images, so for completeness, we also compute reconstructions and quantitative results for this setting (included in the supplementary materials). These reconstructions perform strictly worse than those computed from the lightbox setup. We also show a qualitative comparison of the geometry and materials in Figure~\ref{fig:capture}, of each technique in its default setting, where ours are clearly better than the alternative methods. Table~\ref{tbl:abl_config} shows that our ALP reconstruction pipelines give us better results than using current state-of-art inverse rendering methods to get our ALP models.

\begin{figure*}[t]
\centering
\includegraphics[width=\linewidth]{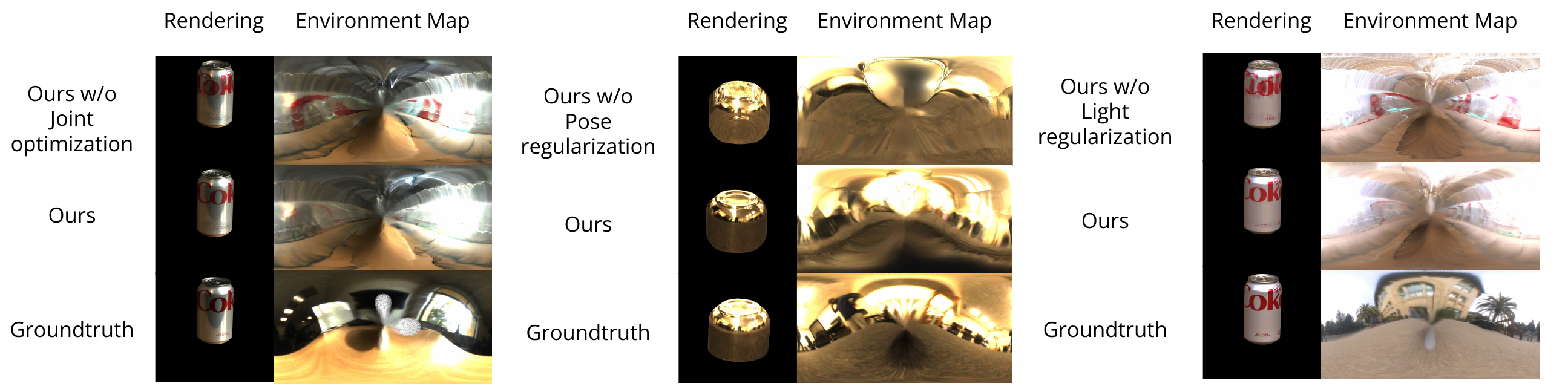}
\vspace{-0.15in}
\caption{Qualitative ablation of the losses we use in our method. Each of our design choices contributes to improvements in pose and lighting optimization which can be observed qualitatively. }
\label{fig:ablation}
\end{figure*}

\myparagraph{Ablation for 6D Pose + Scale Estimation.}
As mentioned in Sec.~\ref{sec:losses}, the pose estimation problem for aligning a 3D model of an ALP and its appearance in a real image is challenging. The appearance of the object in the real image depends on both its pose and lighting; trying to jointly optimize these can introduce potential failure cases. 
In Sec~\ref{sec:losses}, we describe several design choices \wrt the optimization and loss which address some of these failures cases. In Table~\ref{tbl:abl_pose}, we perform an ablation study on each of these decisions and report quantitative results. We show that all design decisions (Silhouette loss, Chamfer loss~\cite{balan2007detailed}, joint optimization, pose, and light regularization) contribute to the final overall performance. We also show representative examples in Figure~\ref{fig:ablation}. They demonstrate how each design choice helps the pose estimation,  which in return helps lighting estimation. In our supplementary material, we further showcase accurate estimations even under extreme object poses.

\begin{figure}[t]
\centering
\includegraphics[width=1\linewidth]{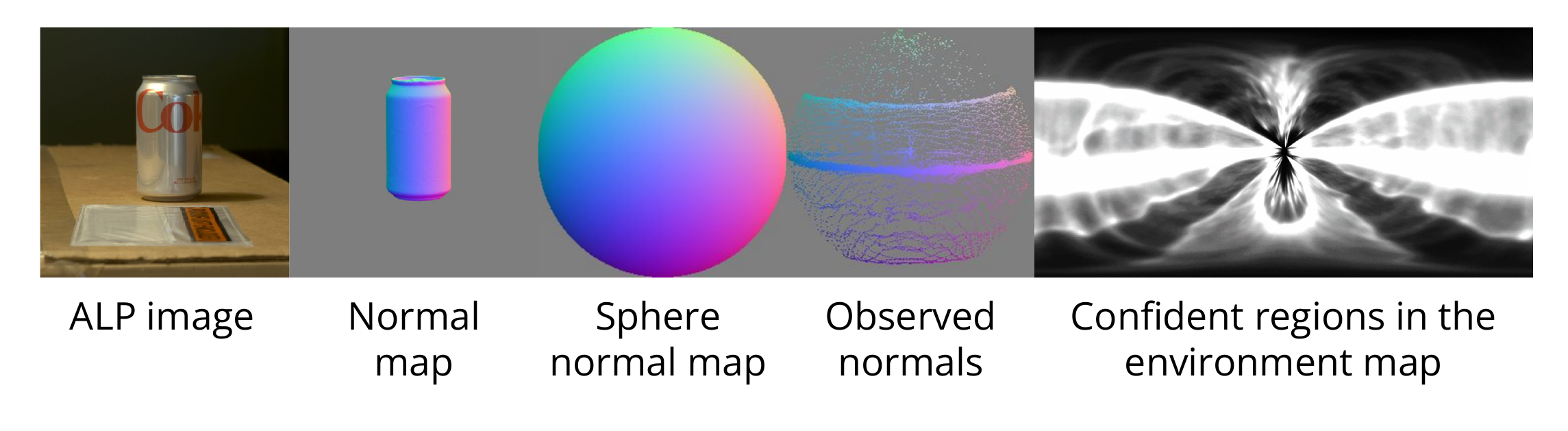}
\vspace{-0.15in}
\captionof{figure}{Visualization of sampling directions for a diet Coke can. See the text in \ref{sec:confidence} for a full description of these visualizations.}
\label{fig:confidence}
\end{figure}

\myparagraph{Visualizing confident regions for \alps.}\label{sec:confidence}
As briefly discussed in Sec.~\ref{sec:light}, an \alp~has a subset of surface normals compared to a perfect sphere light probe, which leads to under-sampled lighting directions. For example, cylindrical objects (Diet Coke can, ring, etc.) tend to sample well light rays perpendicular to the can while significantly under-sampling light rays above and below the can. Since we use VNDF importance sampling which aligns well with our BRDF's density lobe, we visualize a ``confidence map'' as normalized sampling frequency. We show this confidence map in Figure~\ref{fig:confidence} for a representative ALP (i.e., Diet Coke). This demonstrates that the visible surface of a Coke can from a single view only under-samples lighting directions from the top and the bottom.

In our supplementary material, we further show a controlled qualitative analysis of ALPs with different reflectance or shapes to demonstrate that our approach is tolerant to insignificant reflectance and shape variations.

\myparagraph{Discussion.}\label{sec:discussions} 
Our method shows strong promise for recovering scene lighting from a single image containing an ALP. One exciting potential application is improved image editing for in-the-wild images; however, to enable this for \emph{any} image, we would either need to increase the number of ALPs or explore methods that enable us to dynamically edit one of the collected measurements (geometry or material). Another limitation is that we assume our input is an HDR image. However, we note that recent work has sought to convert LDR images to HDR~\cite{lee2018deep,kim2019deep}, and HDR images have become more ubiquitous since many commercial mobile phones now support HDR capture.

 \ignore{If we had an object with a known shape and a prior for the material, then we could solve a constrained three-way optimization for material, lighting, and pose; this would allow us to automatically handle soda cans with different designs.}

\section{Conclusion}

In this paper, we introduced the use of accidental
light probes to estimate environmental lighting from single images.
We did this by first scanning common 3D objects and reconstructing their
reflective properties.  We then used differentiable rendering with a physically-based model to recover the unknown object pose and environment lighting when the
object was placed (or naturally occurred) in an image.
We created a new dataset of materials and geometry for several common,
shiny, curved objects along with images showing these
in a variety of indoor and outdoor environments.
We demonstrate that our approach strongly outperforms previous
approaches in realism and fidelity.

\myparagraph{Acknowledgements.} We would like to thank William T.\ Freeman for the invaluable discussion and for the photo credit, Varun Jampani for helping us with data collection, and Henrique Weber and Jean-Fran\c{c}ois Lalonde for running their methods as comparisons for us. The work was done in part when Hong-Xing Yu was a student researcher at Google and has been supported by gift funding and GCP credits from Google and Qualcomm.

{\small
\bibliographystyle{ieee_fullname}
\bibliography{reference}
}

\end{document}